\documentclass[conference]{IEEEtran}
\usepackage{cite}
\usepackage{amsmath,amssymb,amsfonts}
\usepackage{algorithmic}
\usepackage{graphicx}
\usepackage{textcomp}
\usepackage{xcolor}
\usepackage[hyphens]{url}
\usepackage{multirow}
\usepackage{array}
\usepackage{float}
\usepackage{balance}
\usepackage{svg}
\usepackage{subcaption}
\usepackage{soul}  
\usepackage{comment}
\usepackage{lineno}
\usepackage{mathtools}
\graphicspath{{./figures/}}
\DeclareGraphicsExtensions{.eps,.png,.svg}
\usepackage{fixltx2e}
\usepackage{stfloats}
\usepackage{lscape}
\usepackage[linesnumbered, ruled, vlined]{algorithm2e}

\def\BibTeX{{\rm B\kern-.05em{\sc i\kern-.025em b}\kern-.08em
    T\kern-.1667em\lower.7ex\hbox{E}\kern-.125emX}}

\title{An Adaptive Sampling and Edge Detection Approach for Encoding Static Images for Spiking Neural Networks}
\author{
    \IEEEauthorblockN{Peyton Chandarana\IEEEauthorrefmark{1}\IEEEauthorrefmark{3}, Junlin Ou\IEEEauthorrefmark{2}\IEEEauthorrefmark{3}, Ramtin Zand\IEEEauthorrefmark{1}}
    \IEEEauthorblockA{\IEEEauthorrefmark{1}Department of Computer Science and Engineering, University of South Carolina, Columbia, SC}
    \IEEEauthorblockA{\IEEEauthorrefmark{2}Department of Mechanical Engineering, University of South Carolina, Columbia, SC}
    \IEEEauthorblockA{\IEEEauthorrefmark{3} Authors contributed equally.}
}

\begin{document}
\maketitle

\thispagestyle{plain}
\pagestyle{plain}

\begin{abstract}
Current state-of-the-art methods of image classification using convolutional neural networks are often constrained by both latency and power consumption. This places a limit on the devices, particularly low-power edge devices, that can employ these methods. Spiking neural networks (SNNs) are considered to be the third generation of artificial neural networks which aim to address these latency and power constraints by taking inspiration from biological neuronal communication processes. Before data such as images can be input into an SNN, however, they must be first encoded into spike trains. Herein, we propose a method for encoding static images into temporal spike trains using edge detection and an adaptive signal sampling method for use in SNNs. The edge detection process consists of first performing Canny edge detection on the 2D static images and then converting the edge detected images into two X and Y signals using an image-to-signal conversion method. The adaptive signaling approach consists of sampling the signals such that the signals maintain enough detail and are sensitive to abrupt changes in the signal. Temporal encoding mechanisms such as threshold-based representation (TBR) and step-forward (SF) are then able to be used to convert the sampled signals into spike trains. We use various error and indicator metrics to optimize and evaluate the efficiency and precision of the proposed image encoding approach. Comparison results between the original and reconstructed signals from spike trains generated using edge-detection and adaptive temporal encoding mechanism exhibit 18$\times$ and 7$\times$ reduction in average root mean square error (RMSE) compared to the conventional SF and TBR encoding, respectively, while used for encoding MNIST dataset.
\end{abstract}
\section{Introduction}

Neural encoding plays an important role in the brain-inspired neuromorphic systems, which use trains of action potentials, \textit{i.e.} spike trains, to process the input information. Neural encoding refers to the process of representing stimulus features in the form of spike trains. Before any type of learning or training can occur in spiking neural networks (SNNs), these spike trains must first be produced from input data using particular encoding mechanisms. As also mentioned in \cite{neural_encoding}, the study of neural coding can be divided into three questions, what is being encoded? how is it being encoded? and what is the precision of the encoding?

In this paper, we are focusing on the encoding of static images for image classification with SNNs into spike trains. Two of the well-known methods for encoding static images are firing rate-based encoding \cite{ratecoding1,ratecoding2,ratecoding3} and population rank order encoding \cite{rankorder1,rankorder2}. In the rate-based encoding, each input is a Poisson spike train with a firing rate proportional to the intensity of the of the corresponding pixel in the image \cite{ratecoding3}. Although this approach has shown to be effective in practice, it normally requires very high firing rates, therefore it is not computationally efficient \cite{Rueckauer2017, Pfeiffer2018}. The population rank order encoding utilizes receptive fields to distribute input over several neurons, each of which fires only once during the coding interval based on the extent an input value belongs to its corresponding receptive field \cite{selectionencoding, LOBO20181}. This method drastically reduces the spikes that should be processed by network, however it significantly increases the network size. Another common approach is to represent the pixels of the image themselves as spikes and represent each pixel as a separate neuron in the input layer of the SNN \cite{vaila2019deep}. This approach, however, requires a large network of neurons compared to the even the aforementioned approaches. Even with some feature reduction in the images the network is still non-ideal in its size and immutable input size \cite{vaila2019deep}. Herein, we use temporal coding, which is traditionally suitable for encoding time-series and streaming data, to encode static images. Temporal encoding is an efficient and fast method that generates spikes at exact points in time based on the change in the input stimulus signal.

To encode static images into the signals needed as input to the previously mentioned encoding algorithms, we use edge detection along with a conversion algorithm to convert the static images into two signals which will then be converted into two separate spike trains to be input into an SNN. As mentioned above, one of the important questions that needs to be addressed in the neural coding is the precision of the encoding \cite{neural_encoding}. The objective here is to generate spike trains with reduced number of spikes without losing the important information content of the signal. Therefore, evaluating the efficiency of the temporal encoding mechanism becomes very important. In this paper, in order to assess our proposed approach, we use MNIST handwritten digits \cite{MNIST} as representative static images, convert them to signals, encode the signal into spike trains, reconstruct the signals and compare them with original signals, and eventually reconstruct the static MNIST images.   

The remainder of this work is organized as follows. Section 2 discusses the image-to-signal conversion approach using an edge detection algorithm. Section 3 describes the adaptive temporal encoding mechanism. Section 4 provides a thorough evaluation of our proposed static image encoding method using intensive simulations and introducing a new fitness function. Finally, section 5 concludes the paper by proposing possible directions for future work.
\begin{figure}
\centering
\includegraphics[width=3.3in]{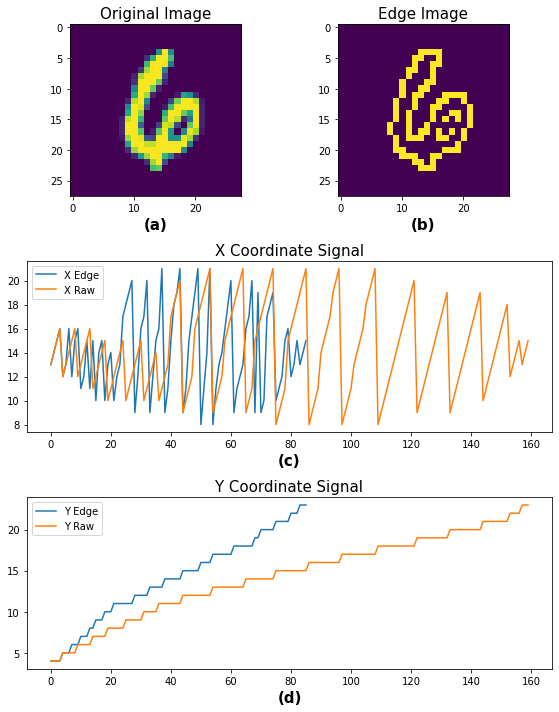}
\caption{(a) The original image. (b) Canny edge detection output. (c) X coordinate signal from image (b). (d) Y coordinate signal from image (b). Algorithm \ref{algo:coord_extract} was used to obtain the signals in (c) and (d).}
\label{fig:x_y_edge_image}
\end{figure}

\begin{algorithm}
\DontPrintSemicolon
\SetNlSty{text}{}{:}
\SetAlgoNlRelativeSize{0}
\SetKwData{Thresh}{threshold}\SetKwData{X}{x}\SetKwData{Y}{y}\SetKwData{Image}{image}
\SetKwData{List}{[\ ]}
\SetKwFunction{Append}{.append}\SetKwFunction{Len}{length}
\KwIn{A 2D image array of pixel intensities}
\KwOut{Two arrays, x \& y, consisting of the x and y components of pixels with intensities greater than zero}
\BlankLine
\X$= \List$\;
\Y$= \List$\;
\For{$row = 1$ \KwTo $\Len(\Image)$}{
 \For{$col = 1$ \KwTo $\Len(\Image[row])$}{
  \If{$\Image[row][col] > 0$}{
   $\X\Append(col)$\;
   $\Y\Append(row)$\;
  }
 }
}
\caption{Coordinate Extraction}
\label{algo:coord_extract}
\end{algorithm}
\section{Image-to-Signal Conversion Method}
In image processing, edge detection is used to find the edges of objects or shapes in images. This process, in general, consists of identifying where there is an abrupt change in the pattern of pixel intensities. The edge of an object represents the object's general shape or contour and therefore aids in identifying that object. This idea that an object's edges may contain enough information to classify an image, is what led us to investigate if encoding the edges of an image instead of the image itself would lead to a reduction in spikes in spike trains. 

From here, we implemented a Python program which employs the OpenCV Canny edge detection to find the edges of the MNIST handwritten digit dataset \cite{opencv_library}. Canny edge detection consists of four different image processing steps: (1) Gaussian Filter, (2) Sobel Filters, (3) Non-maximum Suppression, and (4) Hysteresis Thresholding.

The first two steps in the Canny edge detection process involve applying two filters to the image using convolution. Convolution allows neighboring pixels around a central input pixel to be weighted and then summed up to determine the intensity of the central output pixel in the same location as the input pixel. In the case of Canny edge detection, a 5x5 Gaussian filter is first applied to smoothen the image and reduce the image noise \cite{opencv_library}.

The Sobel filter is then applied to the smoothened image to get a preliminary edge detected image. Applying the Sobel filter computes the first derivative in both the horizontal and vertical directions which corresponds to two output images. These two output images, horizontal and vertical, are then used to find the edge gradient and the direction of the line perpendicular to the tangent line of the edge \cite{opencv_library}. The result is an image which has pixels with higher intensities roughly where the edges were located.

While the resulting image from applying the Sobel filter process gives us the edges of the image, these edges are not ideal and the number of edge pixels are not minimized. This is not ideal because we wish to minimize the number of pixels that represent the edges so that the signal length and consequently spike train length is minimized. The next steps in Canny edge detection aid in minimizing the edge pixels. In the third step, a non-maximum suppression algorithm is used to suppress certain pixels that are not actually part of an edge. The algorithm looks at the pixels in the output of the second step and then, using the direction from the second step, analyzes the pixel and its surrounding pixels to determine if it is on the edge. If the pixel is determined to not be on the edge, then its intensity is set to zero \cite{opencv_library}.

The final step in the Canny edge detection is to use hysteresis thresholding to eliminate any remaining pixels which are not actual edge pixels and output an image with semi-connected thin edges. The algorithm, specifically, looks for pixels which have intensities higher than a given threshold and determines them as edge pixels. If the pixel has an intensity smaller than a given lower threshold, then it is determined to not be on the edge. Finally, if the pixel is between the upper and lower intensity thresholds, then pixel is determined to be on the edge if and only if it is connected to an already determined edge pixel \cite{opencv_library}.

After applying the Canny edge detection process, the edges of the original image in Figure \ref{fig:x_y_edge_image} (a) are detected and Figure \ref{fig:x_y_edge_image} (b) is generated. Figure \ref{fig:x_y_edge_image} (b) is then passed into Algorithm \ref{algo:coord_extract} to extract the X and Y coordinates of the edge pixels in the image and create the X and Y signals which will be later encoded into spike trains. As shown in Figure \ref{fig:x_y_edge_image} (c) and (d), there is a significant reduction in the length of the X and Y signals obtained from the edge image compared to those of the original raw image. In particular, for a random sample of 1000 MNIST digits, \textit{i.e.} 100 samples of each digit 0 through 9, there is a 41.7\% reduction in the length of the signal for edge images compared to the original images.
\begin{algorithm}
\DontPrintSemicolon
\SetNlSty{text}{}{:}
\SetAlgoNlRelativeSize{0}
\SetKwData{Thresha}{$SamplingThreshold$}\SetKwData{Threshb}{$EncodingThreshold$}\SetKwData{S}{s}\SetKwData{Count}{count}
\SetKwData{SA}{$s_a$}\SetKwData{N}{n}\SetKwData{Out}{out}\SetKwData{SP}{startpoint}
\SetKwFunction{Len}{length}\SetKwFunction{Ceil}{ceil}
\SetKwFunction{Zeros}{zeros}\SetKwFunction{Abs}{abs}\SetKwFunction{Sum}{sum}
\SetKwFunction{TEnc}{TemporalEncoding}
\KwIn{\S signal, \Thresha, \Threshb}
\KwOut{\Out, \SP, \Count}
\BlankLine
\Count$= \Zeros(\Len(\S)-1)$\;
\For{$t = 0$ \KwTo $\Len(\S)-1$}{
    $\Count[t] = \Ceil(\Abs((\S[t+1]-\S[t]) / \Thresha))$\;
}
\SA$= \Zeros(\Sum(\Count)+1)$\;
\N$= 0$\;
\For{$i = 0$ \KwTo $\Len(\S)-1$}{
 \For{$j = 0$ \KwTo $\Count[t]$}{
  $\SA[\N] = \S[i] + (j) / \Count[t]*(\S[i+1] - \S[i])$\;
  \N$+=1$\;
 }
}
$\SA[\Len(\SA)-1] = \S[\Len(\S)-1]$\;
$\Out, \SP = \TEnc(\SA, \Threshb, ...)$\;
\caption{Temporal Encoding with Adaptive Sampling}
\label{algo:adapt_encode}
\end{algorithm}
\begin{algorithm}
\DontPrintSemicolon
\SetNlSty{text}{}{:}
\SetAlgoNlRelativeSize{0}
\SetKwData{Threshb}{$EncodingThreshold$}\SetKwData{S}{s}\SetKwData{Count}{count}
\SetKwData{SA}{$s_a$}\SetKwData{Out}{out}\SetKwData{SP}{startpoint}\SetKwData{VarM}{m}
\SetKwData{Recon}{$recon$}\SetKwData{Recona}{$recon_a$}\SetKwData{Spikes}{spikes}
\SetKwFunction{Append}{.append}\SetKwFunction{Len}{length}\SetKwFunction{Ceil}{ceil}
\SetKwFunction{Zeros}{zeros}
\SetKwFunction{Dec}{Decoding}
\KwIn{\Spikes, \Threshb, \SP, \Count}
\KwOut{\Recon}
\BlankLine
 \Recona$=\Dec(\Spikes,\Threshb,\SP)$\;
 \VarM$= 0$\;
 \Recon$ =\Zeros(\Len(\Count)+1)$\;
 \For{$t = 0$ \KwTo $\Len(\Count)$}{
  \Recon$[t]=\Recona[\VarM]$\;
  \VarM$+= \Count[t]$\;
 }
 \Recon$[\Len(\Recon)-1]=\Recona[\Len(\Recona)-1]$\;
\caption{Temporal Decoding with Adaptive Sampling}
\label{algo:adapt_decode}
\end{algorithm}
\section{Adaptive Temporal Encoding}
\par
Once the static images are converted to signals, as described in the previous section, temporal encoding mechanisms can be used to generate spike trains from the produced signals. In this step, selection and optimization of temporal encoding methods play an important role to achieve a high precision signal-to-spike train conversion. In our initial experiments, we used the known step-forward (SF) encoding \cite{selectionencoding} method to encode the X and Y signals obtained from edge detection. However, we noticed that using a fixed sampling rate to encode the signal will lead to either an overestimation (with high sampling rate) or an underestimation (with low sampling rate), which leads to high computation cost or low encoding precision, respectively. Thus, in this work, we propose an adaptive sampling approach, which works as a preprocessing step for temporal encoding mechanisms. 

The objective of the adaptive sampling is to increase the sampling rate when there are abrupt changes to the signal and reduce the sampling rate when the signal exhibits more gradual variations. The adaptive sampling algorithm for temporal encoding is presented in Algorithm \ref{algo:adapt_encode}. The algorithm first samples the signal with fixed intervals and then calculates how many more samples, (\textit{\textbf{count}} in Algorithm \ref{algo:adapt_encode}), are required between two consecutive points based on the change in signal between the two points. Next, the signal is sampled again, this time based on the number of samples calculated in the previous step, inserting more points for higher signal resolution based on the rate of change. Finally, the adaptively-sampled signal is converted to a spike train using any temporal encoding mechanisms such as threshold-based representation (TBR), SF, etc. \cite{selectionencoding}, as shown in line 11 of the Algorithm \ref{algo:adapt_encode}. 

The adaptive sampling approach includes a sampling threshold which defines the sensitivity of the algorithm to changes in the signal. The lower the threshold value is, the more samples will be taken (Line 3 of the Algorithm \ref{algo:adapt_encode}) between two consecutive points. This adaptive sampling thus optimizes the average firing rate (AFR) to be seen in section \ref{sec:sim-res} for minimal spikes in the resulting spike trains without compromising accuracy.

Decoding the spike trains generated by adaptive sampling-based temporal encoding approaches requires a post-processing step that is described in Algorithm \ref{algo:adapt_decode}. Once the spike trains are decoded using normal decoding methods for spike trains \cite{selectionencoding} (Line 1 of Algorithm \ref{algo:adapt_decode}), the sample counts stored in the \textbf{\textit{count}} array during the adaptive sampling will be used to reconstruct the original signal. Here, without loss of generality, we used the adaptive sampling approach along with SF and TBR methods to develop an adaptive temporal spike encoding mechanism for static images. The TBR and SF encoding algorithms and their corresponding decoding algorithm is provided in Appendix A.
\begin{figure*}[]
\centering
\includegraphics[width=6.5in]{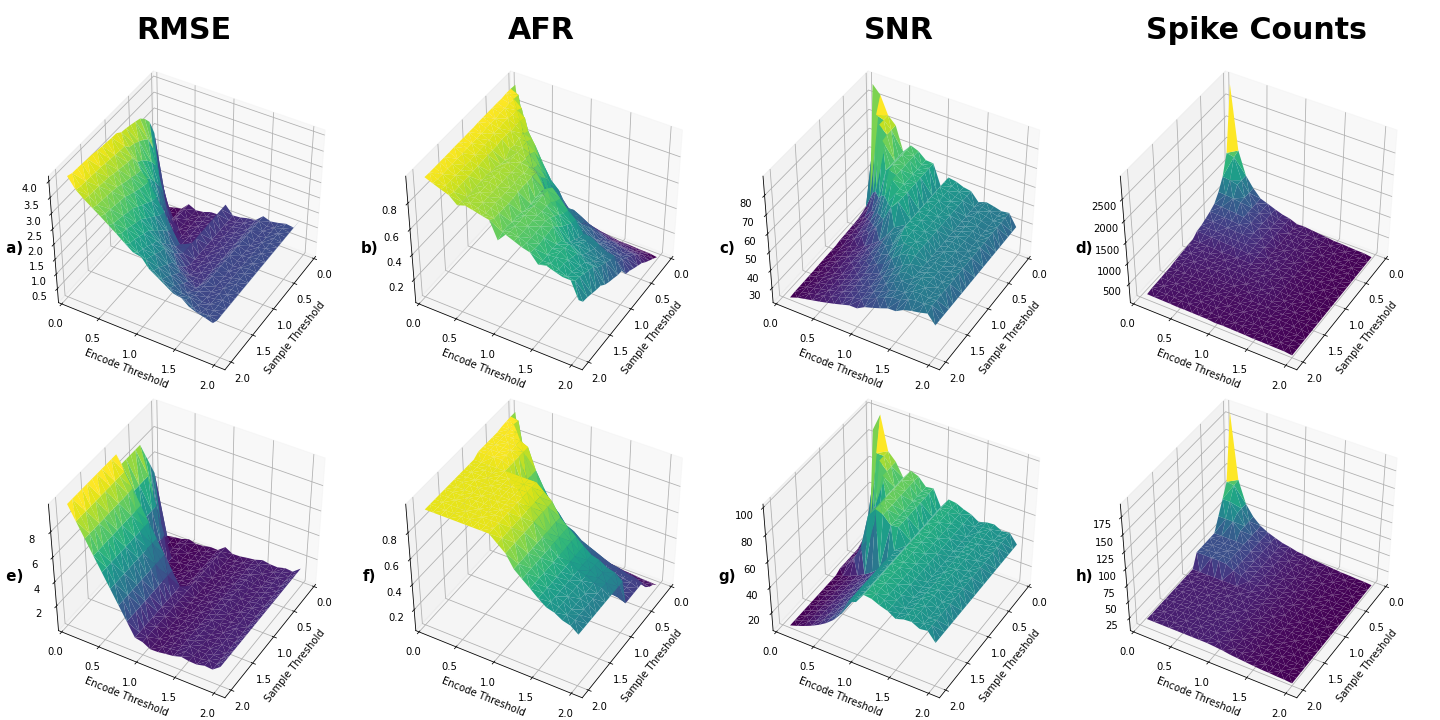}
\caption{Performance metrics for adaptive SF encoding using various  threshold values. (a)-(d) X signal and (e)-(h) Y signal.}
\label{fig:sf_metrics}
\end{figure*}
\begin{figure*}[!t]
\centering
\includegraphics[width=\textwidth]{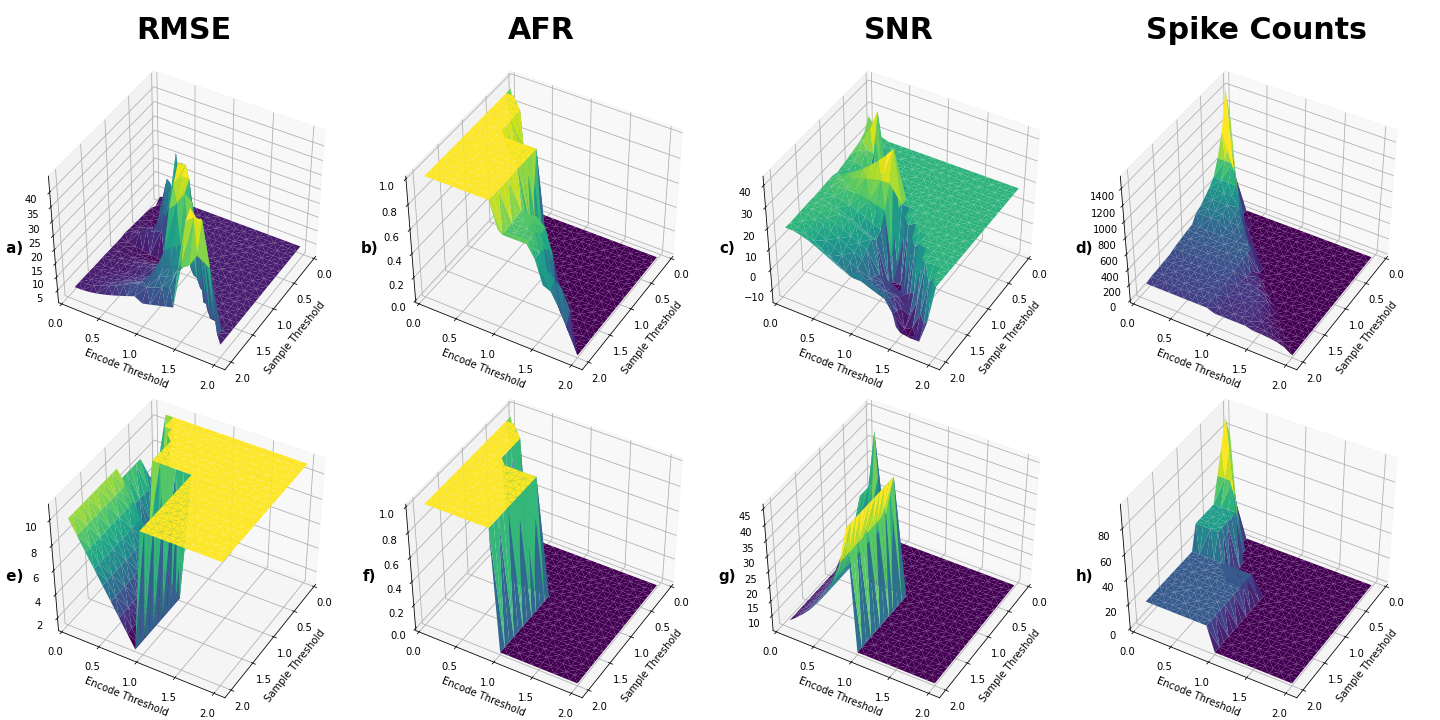}
\caption{Performance metrics for adaptive TBR encoding using various  threshold values. (a)-(d) X signal, and (e)-(h) Y signal.}
\label{fig:tbr_metrics}
\end{figure*}
\section{Simulation Results}
\label{sec:sim-res}
In this section, we evaluate our adaptive temporal encoding approach by using MNIST handwritten digits \cite{MNIST} as representative static images, convert them to signals, encode the signal into spike trains, reconstruct the signals and compare them with original signals using various indicator and error metrics. For samples of complete encoding-to-reconstruction process on MNIST 0-9 digits images, readers can refer to Appendix B.
\subsection{Error and indicator Metrics}
Here, the difference between the original and reconstructed signals is considered to assess the precision and efficiency of the encoding methods. We use root mean square error (RMSE) and signal-to-noise ratio (SNR) metrics to assess the precision of the encoding approaches. SNR is defined as
\begin{equation}
SNR = 20 \cdot \log \frac{Power(s)}{Power(s-r)} [dB]
\end{equation}
where $s$ is the original signal, $r$ is the reconstructed signal, and $Power$ is calculated as
\begin{equation}
    Power (s)=1/N \sum_{k=0}^{N-1}|s(k)|^2
\end{equation}
\par
\noindent where $N$ is the number of samples. RMSE is defined as
\begin{equation}
RMSE = \sqrt{\frac{\sum_{t=1}^N(s_t-r_t)^2}{N}}
\end{equation}
On the other hand, average firing rate (AFR) and spike count metrics are utilized to evaluate the computational efficiency of encoding methods. AFR indicates how saturated the spike train is and is defined as
\begin{equation}
AFR = \frac{\sum_t^N|sp_t|}{N}
\end{equation}
where $sp$ is the number of all spikes in a given spike train \cite{selectionencoding}. Here, the objective is increasing the SNR while decreasing RMSE, AFR and spike count. Figures \ref{fig:sf_metrics} and \ref{fig:tbr_metrics} exhibit the performance metrics for the adaptive and edge detected SF and TBR encoding experiments, respectively, with encoding and sampling thresholds ranging from 0.1 to 2.0 with 0.1 intervals.
\subsection{Threshold Optimization}
\par
\begin{figure}[!t]
\centering
\includegraphics[width=\linewidth]{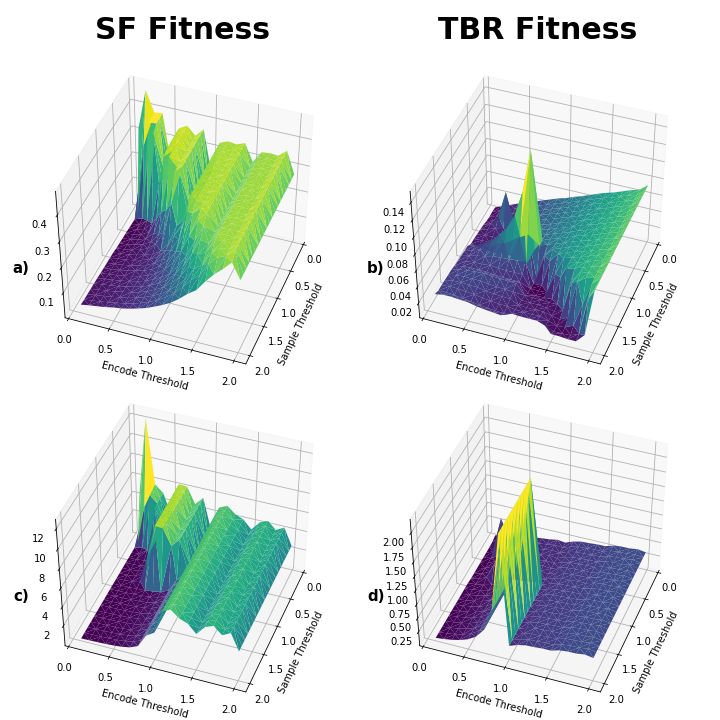}
\caption{Fitness plots. (a)-(b) The X signal. (c)-(d) The Y signal. (a) and (c) use SF encoding. (b) and (d) use TBR encoding.}
\label{fig:sf_tbr_fitness}
\end{figure}

To co-optimize the sampling threshold in the adaptive sampling algorithm and encoding threshold in the temporal encoding mechanisms (see Appendix A), we introduce a Fitness function that combines several of the aforementioned metrics, as expressed in (\ref{alg:fitness}). The higher $Fitness$ value represents better reconstruction.
\begin{equation}
\label{alg:fitness}
    Fitness = \frac{SNR}{{RMSE}\times{Spike Count}}
\end{equation}
Figure \ref{fig:sf_tbr_fitness} shows the reconstruction fitness of the adaptive SF and TBR encoding methods for various sampling and encoding thresholds ranging from 0.1 to 2.0. The maximum fitness values for adaptive SF and TBR encoding methods are achieved at (\textit{SamplingThreshold}, \textit{EncodingThreshold}) equals (0.1, 0.2) for SF and (1.0, 0.9) for TBR, respectively. By contrast, the optimal \textit{EncodingThreshold} for non-adaptively sampled experiments are 2.0 (X) and 1.2 (Y) for SF and 2.0 (X) and 0.9 (Y) for TBR. Table \ref{tab:comparison} provides a comparison between adaptive SF and TBR encoding with the optimized threshold values mentioned above. As listed in the table, TBR is more efficient in terms of spike count, but the SF exhibits a greater SNR and smaller RMSE making it more desireable for precise reconstuction. It should be noted, however, that the average number of spikes is greatly increased due to the adaptive sampling increasing the resolution of the signal. It is worth noting that the fitness function can be adjusted, using $m$ and $n$ parameters in (\ref{eq:fitness_adjust}), based on the specific application requirement to emphasize on the precision or computation efficiency:        

\begin{equation}
\label{eq:fitness_adjust}
    Fitness (m, n) = \frac{SNR}{{RMSE^{(m)}}\times{Spike Count^{(n)}}}
\end{equation}

As can be seen in table \ref{tab:comparison}, the lowest RMSE values occur in the experiments where adaptive sampling was used versus without adaptive sampling. The average number of spikes in all cases where edge detection was used does marginally reduce the number of spikes. 

\begin{table}[]
    \caption{Performance comparison of the adaptive SF and TBR encoding with optimized thresholds for the x and y signals.}
    \label{tab:comparison}
    \centering
\resizebox{\columnwidth}{!}{%
\begin{tabular}{c c c c c c c c c}
    \hline
    Method & Encoding & & RMSE & AFR & SNR & Avg. Spike \# & Fitness \\
    \hline
    \multirow{4}{1.5cm}{Adapive Sampling w/ Edge Detection}  & \multirow{2}{*}{SF} &
    X & \textbf{0.12} & 0.49 & \textbf{84.33} & 1538.7 & \textbf{0.47}\\
    & &
    Y & \textbf{0.08} & 0.5 & \textbf{101.39} & 95.1 & 12.82\\
    & \multirow{2}{*}{TBR} &
    X & 1.26 & 1.0 & 42.92 & 311.98 & 0.14\\
    & & 
    Y & 1.1 & 1.0 & 45.63 & 20.02 & 2.18\\
    \hline
    \multirow{4}{1.5cm}{Adaptive Sampling w/o Edge Detection} & \multirow{2}{*}{SF} &
    X & \textbf{0.12} & 0.49 & \textbf{84.46} & 1555.9 & 0.46\\
    & & 
    Y & \textbf{0.08} & 0.5 & \textbf{102.72} & 93.64 & \textbf{13.27}\\
    & \multirow{2}{*}{TBR} &
    X & 1.0 & 1.0 & 46.78 & 315.67 & 0.18\\
    & & 
    Y & 1.07 & 1.0 & 46.02 & 19.73 & 2.3\\
    \hline
    \multirow{4}{1.5cm}{Only Edge Detection} & \multirow{2}{*}{SF} &
    X & 3.26 & 0.54 & 27.11 & 49.07& 0.17\\
    & & 
    Y & 0.64 & 0.19 & 54.84 & 15.13 & 5.63\\
    & \multirow{2}{*}{TBR} &
    X & 10.43 & 0.47 & 9.28 & 39.75 & 0.05\\
    & & 
    Y & 1.08 & 0.23 & 46.29 & 19.06 & 2.61\\
    \hline
    \multirow{4}{1.5cm}{Conventional} & \multirow{2}{*}{SF} &
    X & 3.01 & 0.43 & 28.55 & 70.43 & 0.13\\
    & & 
    Y & 0.64 & 0.11 & 54.96 & 14.83 & 5.82\\
    & \multirow{2}{*}{TBR} &
    X & 15.85 & 0.15 & -0.03 & 22.02 & 0.03\\
    & & 
    Y & 1.02 & 0.13 & 46.86 & 18.73 & 2.79\\
    \hline
\end{tabular}
}
\end{table}
\section{Conclusion and Future Work}
\label{sec:conclusion}
In this paper, we proposed an adaptive temporal encoding method to convert the static images into spike trains. The proposed encoding approach aims to allow static images to be processed by SNNs with smaller networks, using only two inputs x and y. However, when it comes to temporal encoding, the encoding precision becomes very important. Therefore, we performed intensive simulations to evaluate and optimize our proposed encoding approach. We introduced a novel fitness function which combines the accuracy and efficiency metrics to optimize the hyperparameters of the proposed adaptive temporal encoding. The results obtained exhibit an accurate spike encoding of MNIST handwritten digits as representative static images. Future work includes evaluating the effectiveness of the proposed approach for the whole SNN application through implementing it on neuromorphic hardware, \textit{e.g.} Loihi \cite{Loihi}, for image classification application on a variety of datasets.



\section*{Acknowledgment}
This work is partially supported by an ASPIRE grant from the Office of the Vice President for Research at the University of South Carolina.

\bibliographystyle{IEEEtran}

\bibliography{refs}
\appendices
\section{Temporal Encoding Algorithms}
\label{appendix:algos}
The TBR and SF encoding algorithms and their corresponding decoding algorithm are as follows \cite{selectionencoding}.
\begin{algorithm}
\DontPrintSemicolon
\SetNlSty{text}{}{:}
\SetAlgoNlRelativeSize{0}
\SetKwData{Thresh}{$EncodingThreshold$}\SetKwData{S}{s}
\SetKwData{BA}{$base$}\SetKwData{N}{n}\SetKwData{Out}{out}\SetKwData{SP}{startpoint}
\SetKwFunction{Len}{length}\SetKwFunction{Zeros}{zeros}
\KwIn{\S signal, \Thresh}
\KwOut{\Out, \SP}
\BlankLine
\SP $= \S[0]$\;
\Out $= \Zeros(\Len(\S))$\;
\BA $= \S[0]$\;
\For{$t = 1$ \KwTo $\Len(\S)$}{
    \uIf{$\S[t] > \BA+\Thresh$}{
   $\Out[t] = 1$\;
   $\BA = \BA+\Thresh$\;
    }
    \ElseIf{$\S[t] < \BA-\Thresh$}{
   $\Out[t] = -1$\;
   $\BA = \BA-\Thresh$\;
   }
}
\caption{SF Encoding}
\label{algo:SF_encode}
\end{algorithm}
\begin{algorithm}
\SetNlSty{text}{}{:}
\DontPrintSemicolon
\SetKwData{S}{s}\SetKwData{Thresha}{$EncodingThreshold$}
\SetKwData{SA}{$s_a$}\SetKwData{SP}{startpoint}\SetKwData{Out}{out}
\SetKwFunction{Zeros}{zeros}\SetKwFunction{Len}{len}\SetKwFunction{Std}{std}\SetKwFunction{Mean}{mean}
\KwIn{\S signal, \Thresha}
\KwOut{\Out}
\BlankLine
$\SP = \S[0]$\;
$diff = \Zeros(\Len(\S))$\;
\For{$t = 0$ \KwTo $\Len(\S)-1$}{
    $diff[t] = s(t+1) - s(t)$
}
$diff[\Len(\S)-1] = diff[\Len(\S)-2]$\;
$\Out = \Zeros(\Len(\S))$\;
\For{$t = 0$ \KwTo $\Len(\S)$}{
    \uIf{$diff[t] > \Thresha$}{
   $\Out[t] = 1$\;
    }
    \ElseIf{$diff[t] > -\Thresha$}{
   $\Out[t] = -1$\;
   }
}
\caption{TBR Encoding}
\label{algo:TBR_encode}
\end{algorithm}
\begin{algorithm}
\SetNlSty{text}{}{:}
\DontPrintSemicolon
\SetAlgoNlRelativeSize{0}
\SetKwData{Thresh}{$EncodingThreshold$}\SetKwData{S}{s}\SetKwData{Spikes}{spikes}\SetKwData{Recon}{$recon$}
\SetKwData{BA}{$base$}\SetKwData{N}{n}\SetKwData{Out}{out}\SetKwData{SP}{startpoint}
\SetKwFunction{Len}{length}\SetKwFunction{Zeros}{zeros}
\KwIn{\Spikes, \Thresh, \SP}
\KwOut{\Recon}
\BlankLine
\Recon $= \Zeros(\Len(\Spikes))$\;
\Recon[0] = \SP\;
\For{$t = 1$ \KwTo $\Len(\Spikes)$}{
    \uIf{$\Spikes[t] == 1$}{
   $\Recon[t] = \Recon[t-1]+\Thresh$\;
    }
    \uElseIf{$\Spikes[t] == -1$}{
   $\Recon[t] = \Recon[t-1]-\Thresh$\;
   }
   \Else{
   $\Recon[t] = \Recon[t-1]$\;
   }
}
\caption{Temporal Decoding}
\label{algo:TEMP_decode}
\end{algorithm}
\section{Image Encoding-to-Reconstruction Examples}
\label{appendix:recon}

\begin{figure*}
    \centering
    \begin{tabular}{cc}
    \subfloat[][]{\includegraphics[width=.3\linewidth]{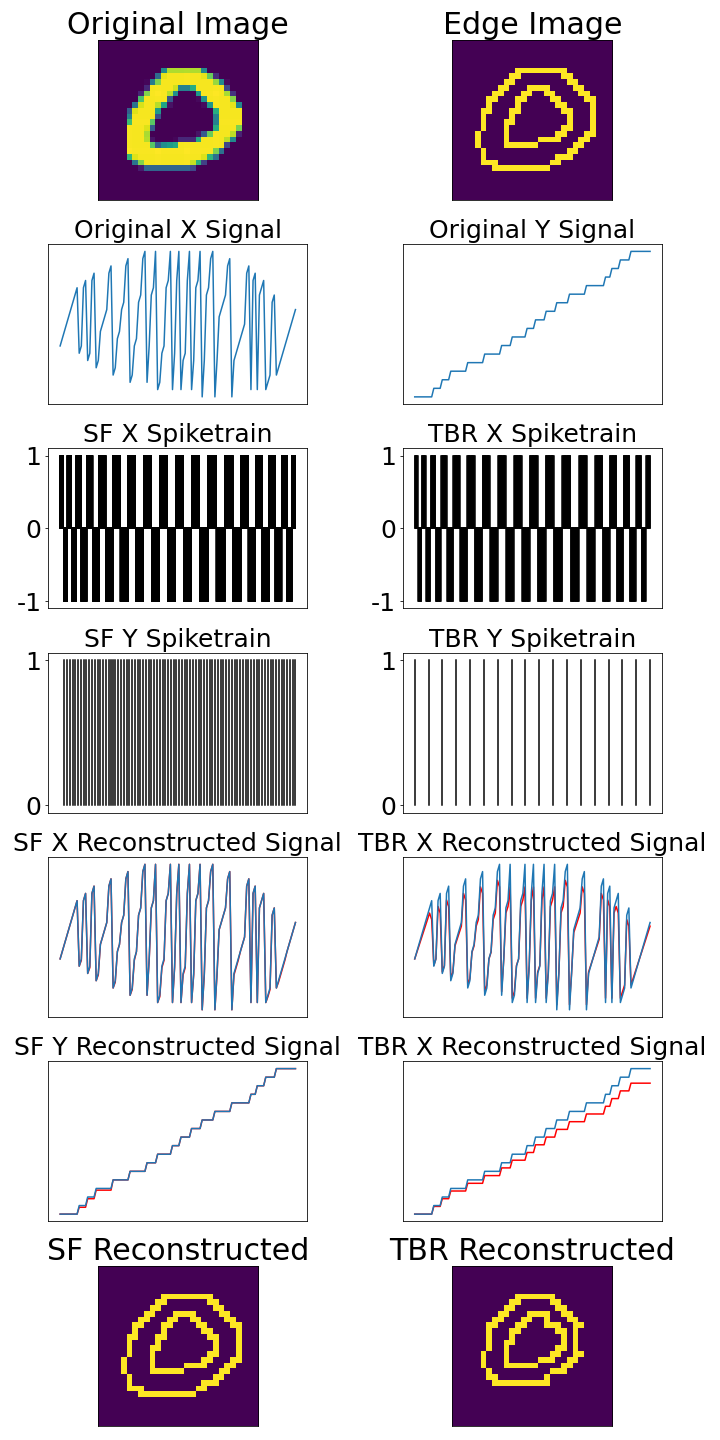}\label{subfig:0}}
    \subfloat[][]{\includegraphics[width=.3\linewidth]{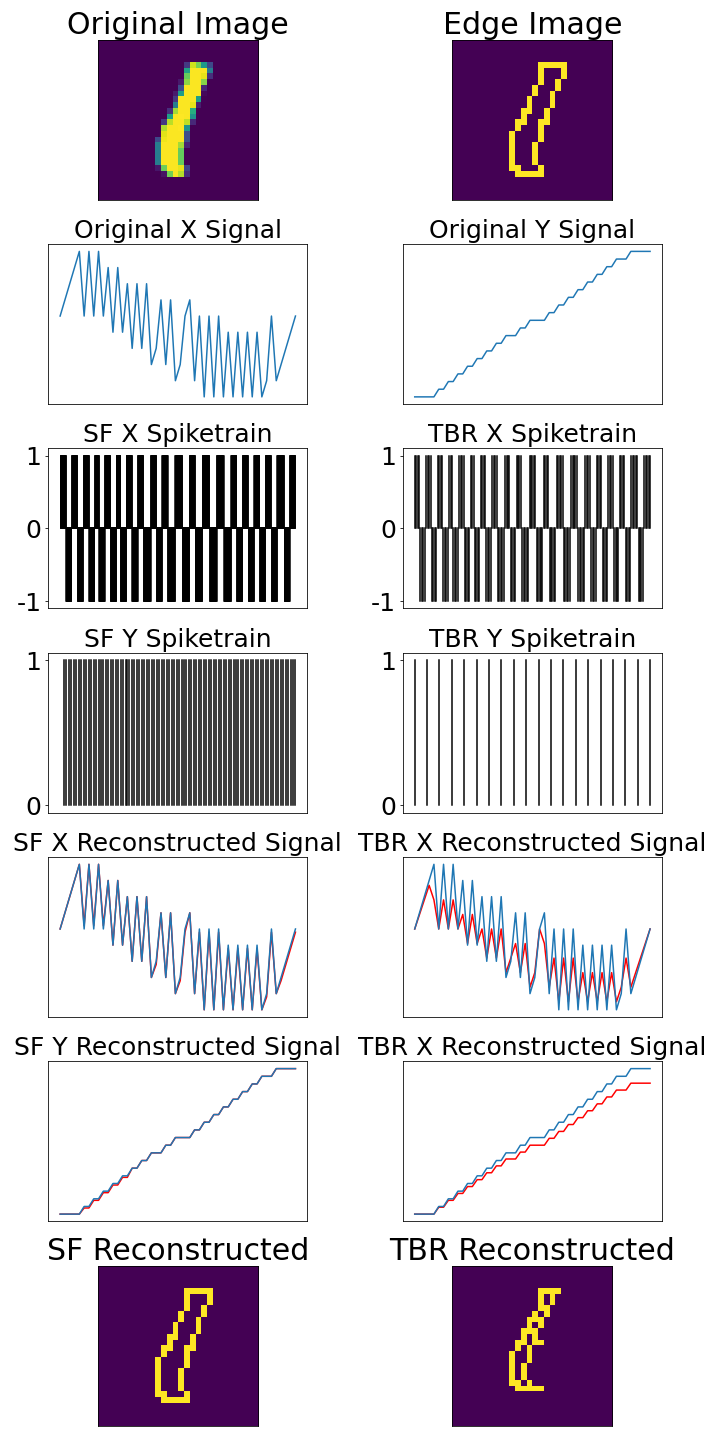}\label{subfig:1}}
    \subfloat[][]{\includegraphics[width=.3\linewidth]{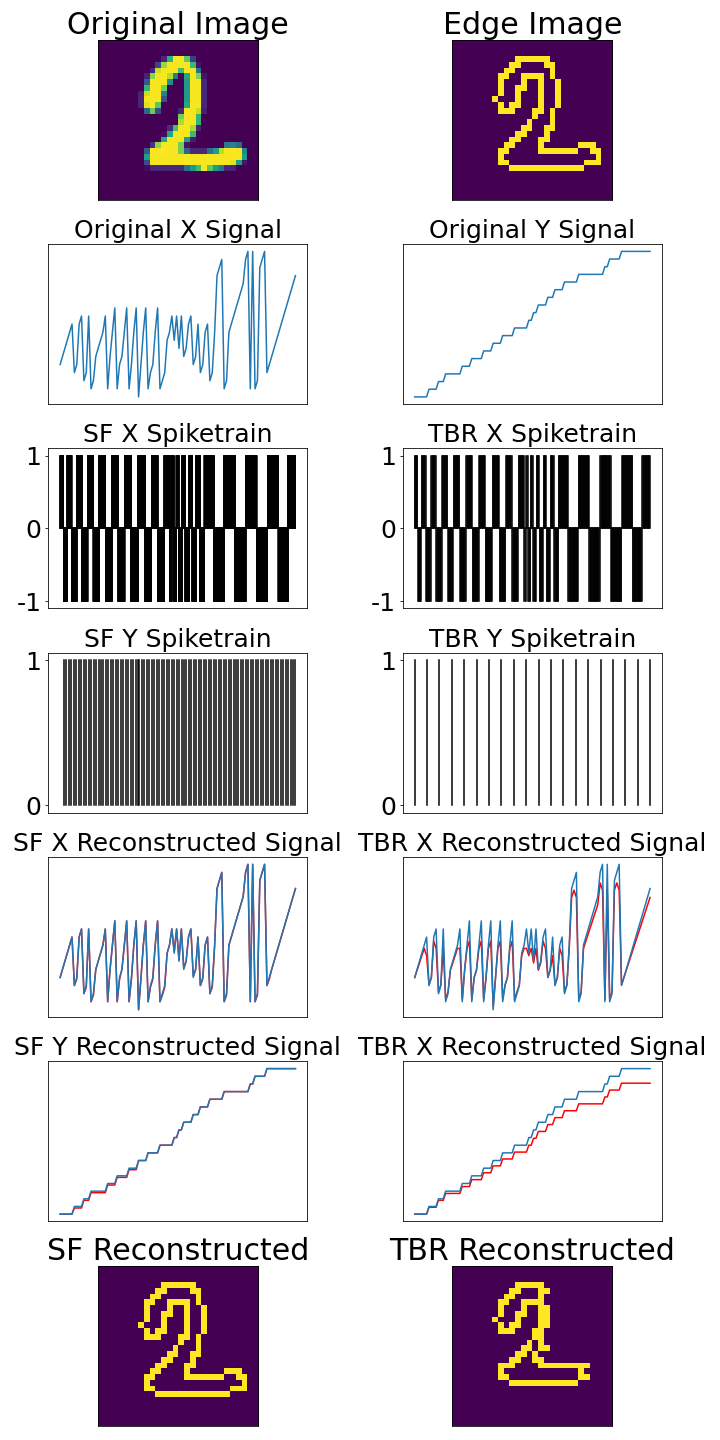}\label{subfig:2}}\\
    \subfloat[][]{\includegraphics[width=.3\linewidth]{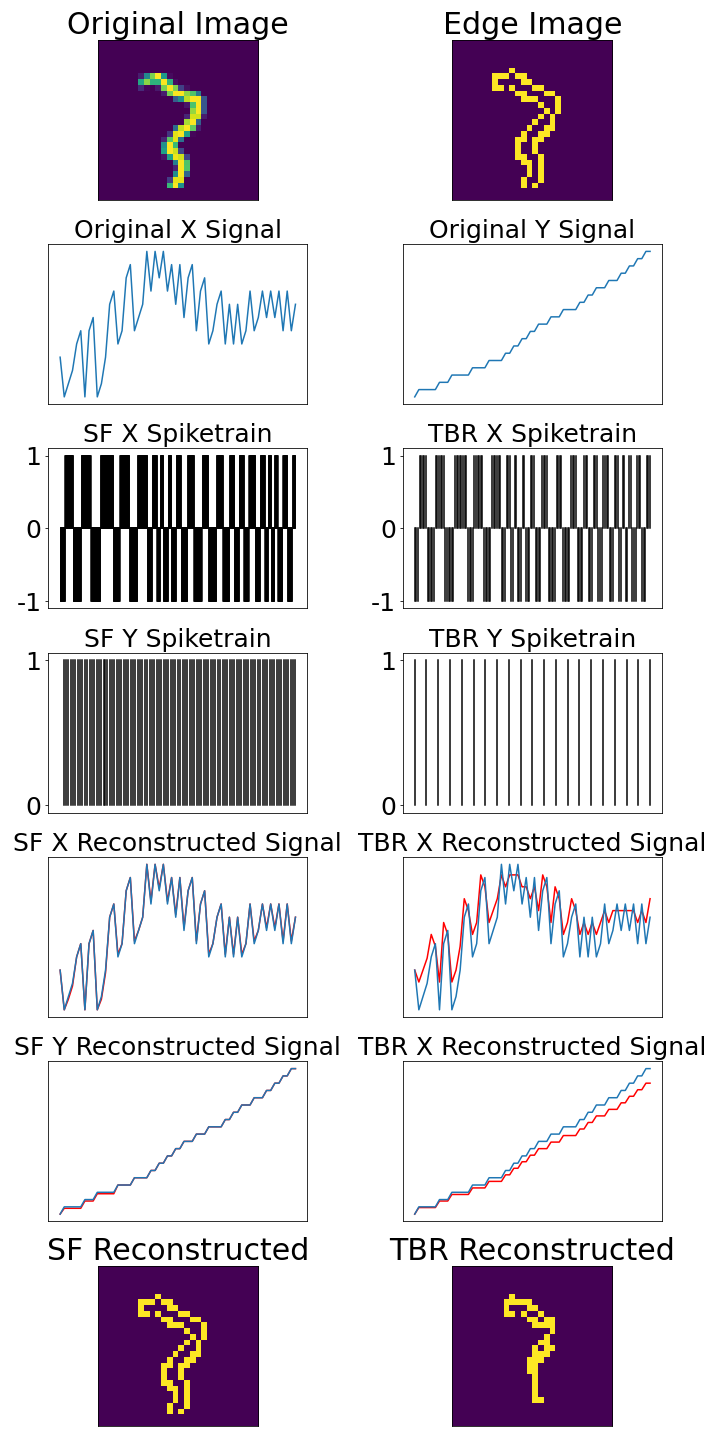}\label{subfig:3}}
    \subfloat[][]{\includegraphics[width=.3\linewidth]{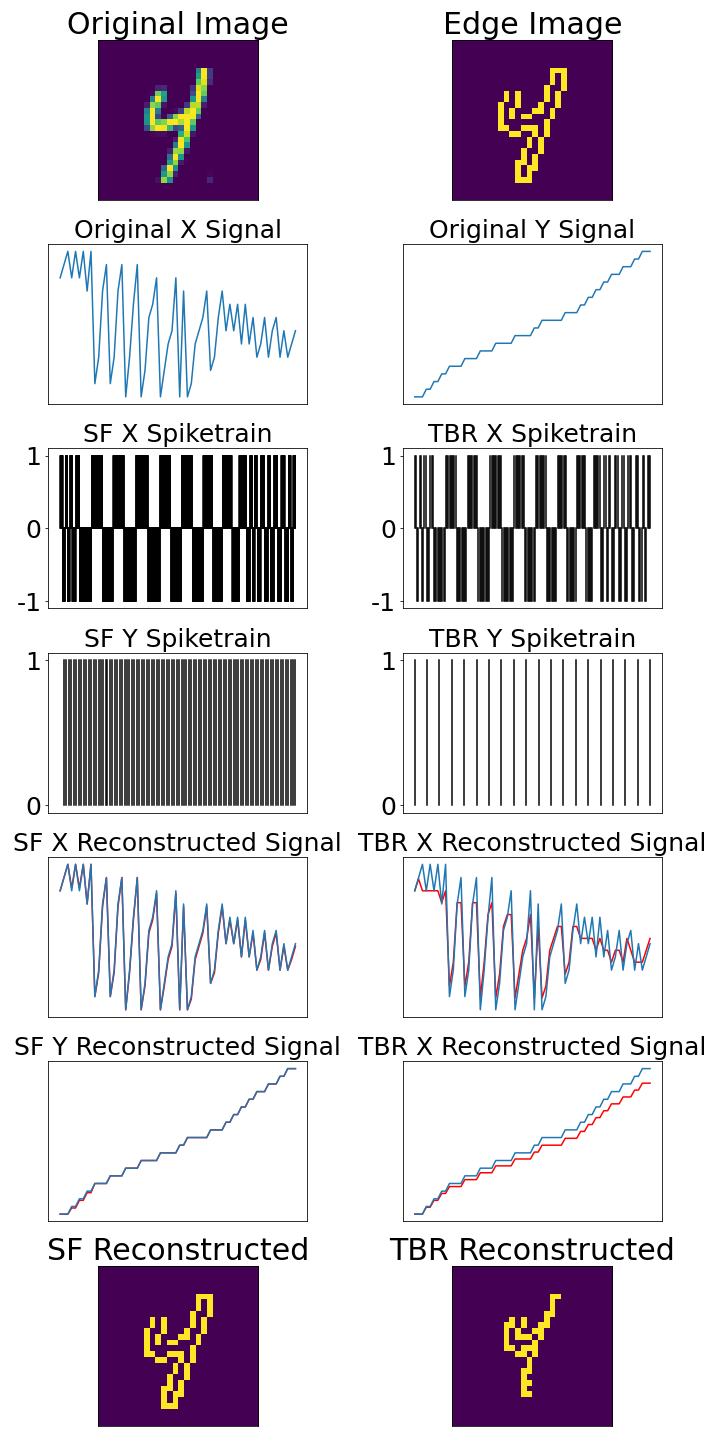}\label{subfig:4}}
    \subfloat[][]{\includegraphics[width=.3\linewidth]{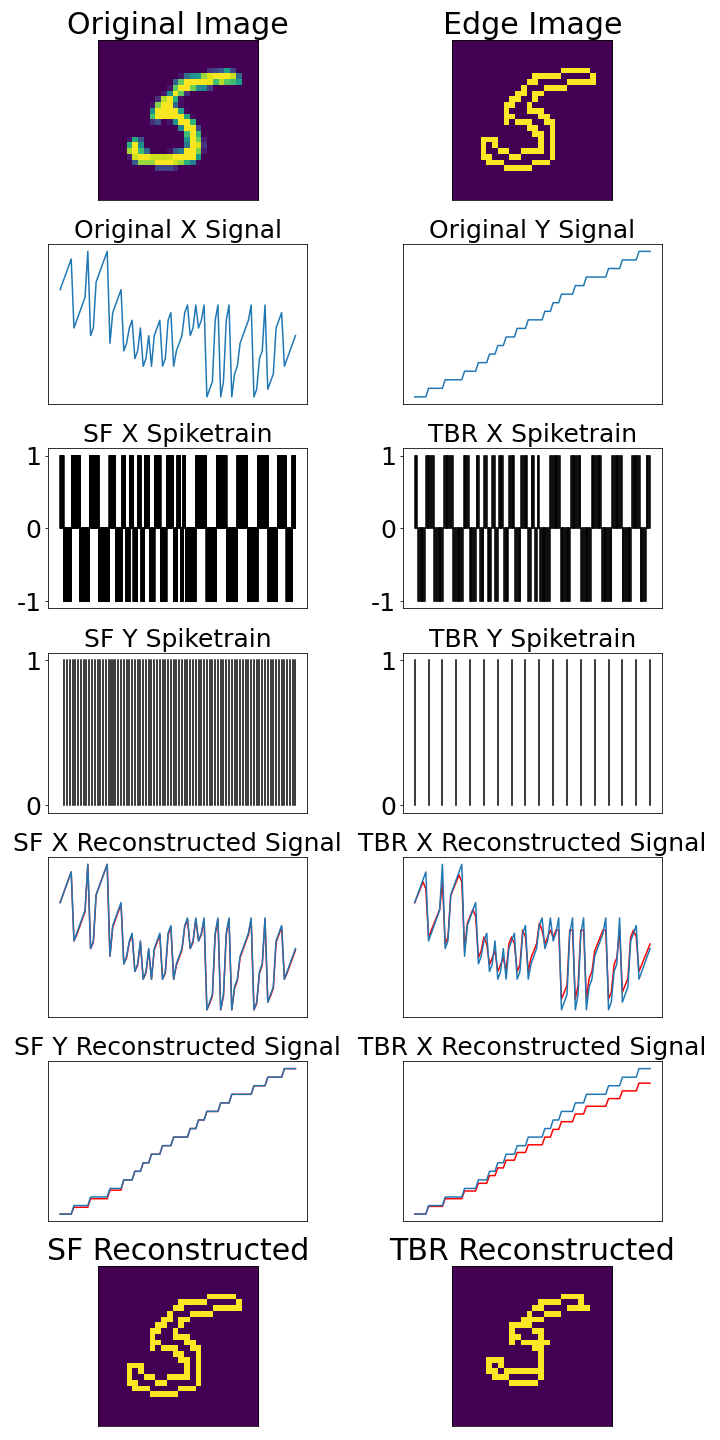}\label{subfig:5}}
    \end{tabular}
    \caption{(a)-(j) Sample of all digits, 0-9, using the optimized sampling and encoding thresholds with SF and TBR encoding. In the reconstructed signal plots, the red and blue lines correspond to the reconstructed and original signals, respectively.}
    \label{fig:my_label}
\end{figure*}

\begin{figure*}
    \ContinuedFloat
    \centering
    \begin{tabular}{cc}
    \subfloat[][]{\includegraphics[width=.3\linewidth]{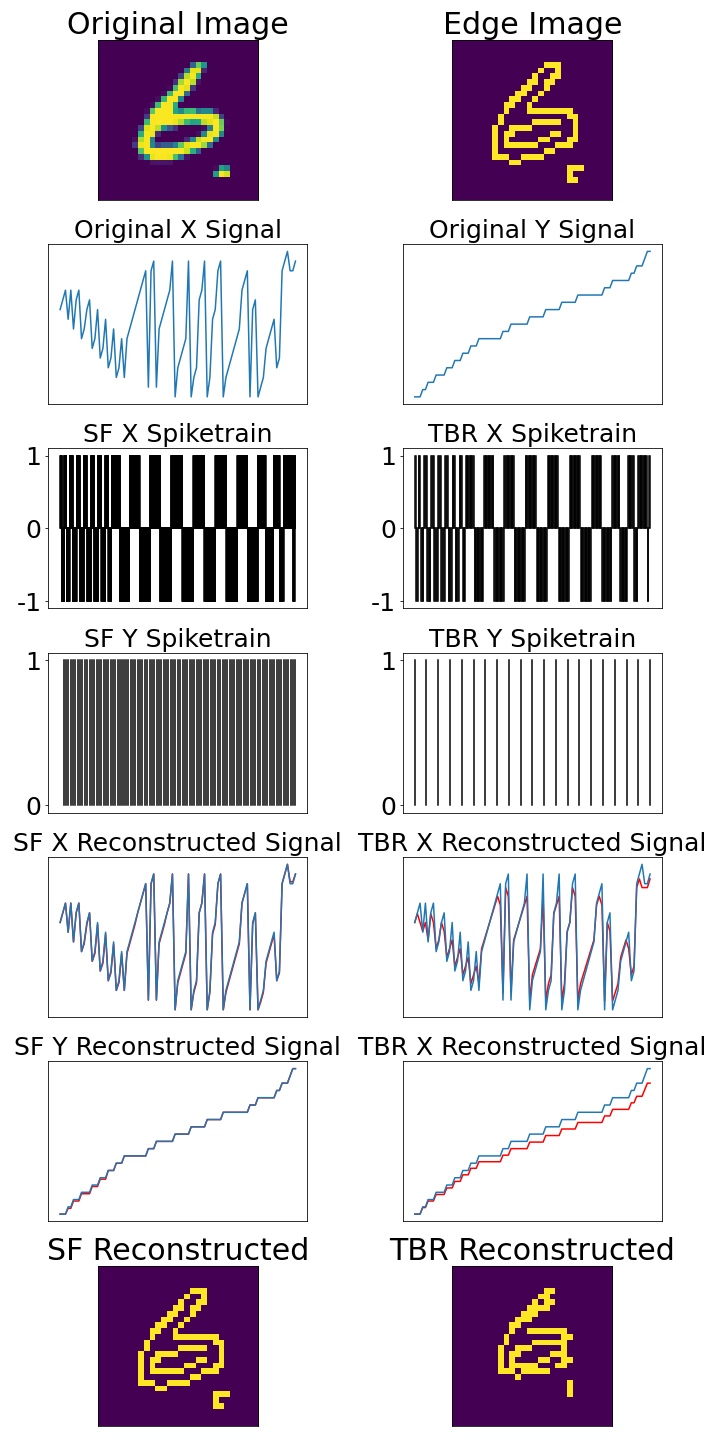}\label{subfig:6}}
    \subfloat[][]{\includegraphics[width=.3\linewidth]{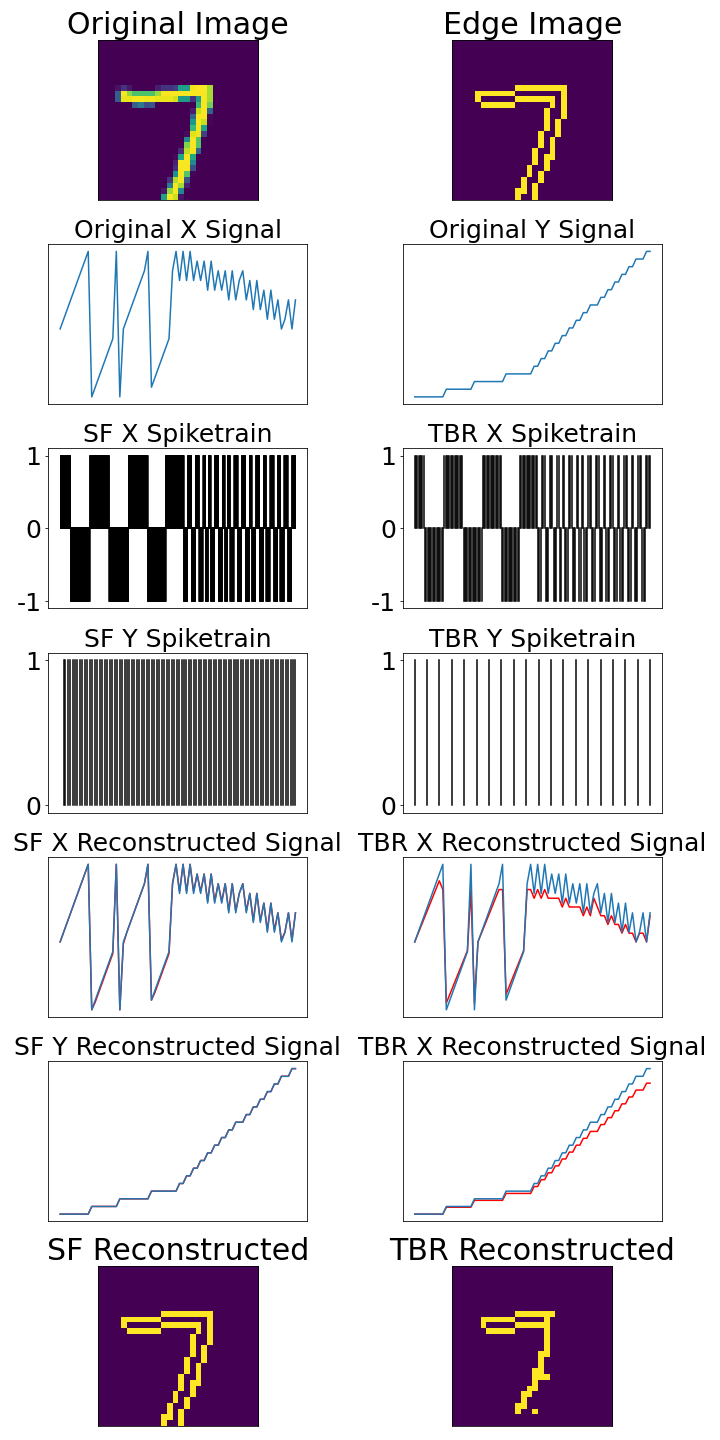}\label{subfig:7}}\\
    \subfloat[][]{\includegraphics[width=.3\linewidth]{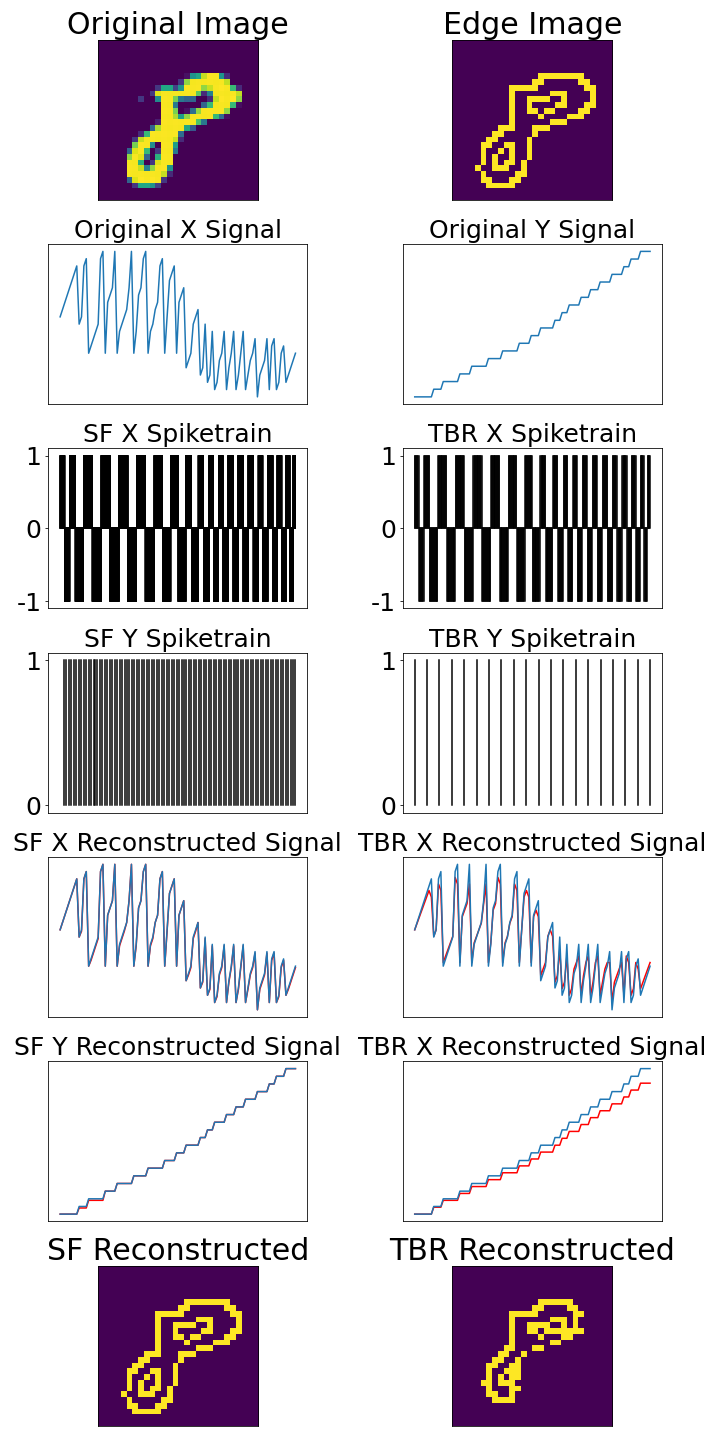}\label{subfig:8}}
    \subfloat[][]{\includegraphics[width=.3\linewidth]{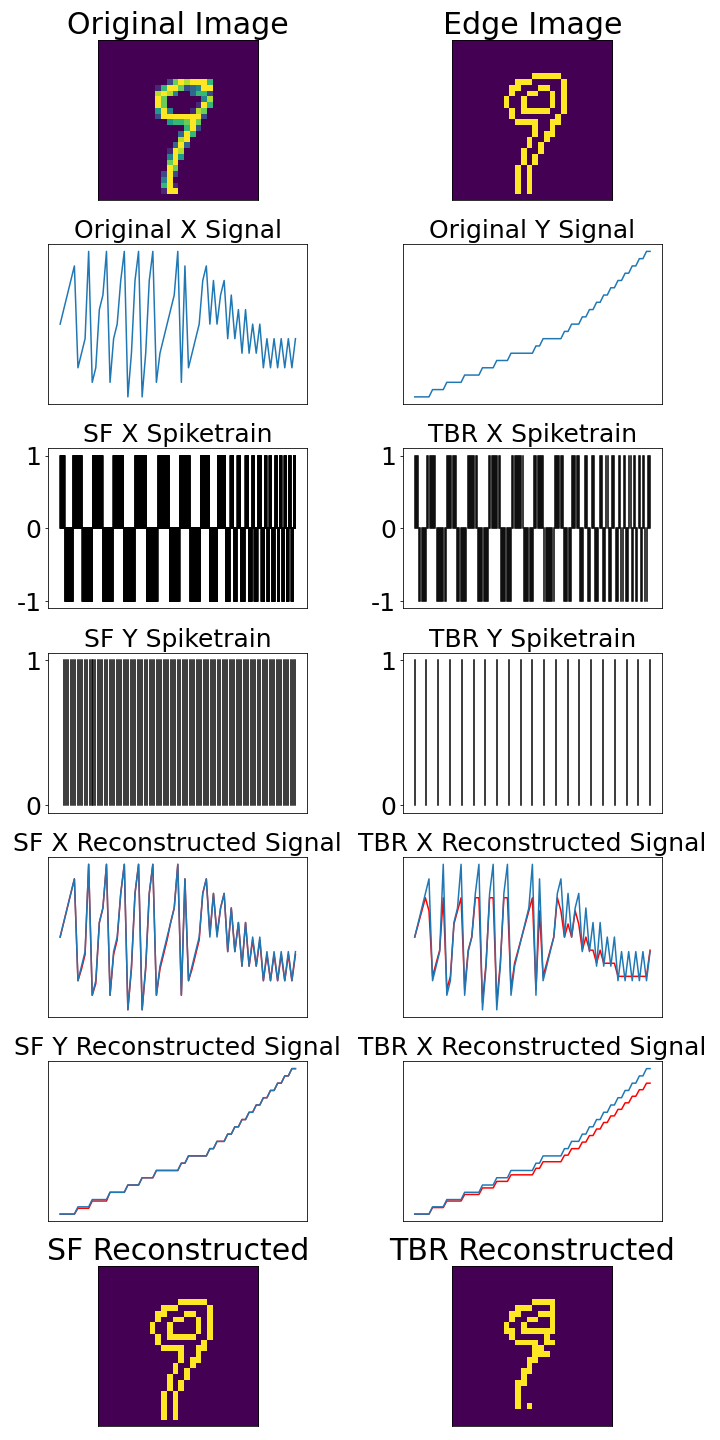}\label{subfig:9}}
    \end{tabular}
    \caption{(a)-(j) Sample of all digits, 0-9, using the optimized sampling and encoding thresholds with SF and TBR Encoding. In the reconstructed signal plots, the red and blue lines correspond to the reconstructed and original signals, respectively.}
    \label{fig:my_label}
\end{figure*}
\end{document}